\documentclass[a4paper,twoside]{article}

\usepackage{epsfig}
\usepackage{subcaption}
\usepackage{calc}
\usepackage{amssymb}
\usepackage{amstext}
\usepackage{amsmath}
\usepackage{amsthm}
\usepackage{multicol}
\usepackage{colortbl}
\usepackage{pslatex}
\usepackage{cleveref}
\usepackage{soul}
\usepackage{apalike}
\usepackage{ wasysym }
\usepackage{listings}
\usepackage{flushend}
\usepackage{booktabs}
\usepackage{todonotes}
\usepackage{SCITEPRESS}     
\newcommand{\mypar}[1]{\noindent\textbf{#1:}}

\soulregister\Cref7
\soulregister{\cite}{7}

\begin{document}

\title{Transfer Learning via Test-Time Neural Networks Aggregation}

\author{\authorname{Bruno Casella\sup{1,2}\orcidAuthor{0000-0002-9513-6087}, Alessio Barbaro Chisari\sup{3,4}\orcidAuthor{0000-0002-7831-382X}, Sebastiano Battiato\sup{4}\orcidAuthor{0000-0001-6127-2470} and Mario Valerio Giuffrida\sup{5}\orcidAuthor{0000-0002-5232-677X}}
\affiliation{\sup{1}Department of Computer Science, University of Torino, Torino, Italy}
\affiliation{\sup{2}Department of Economics and Business, University of Catania, Catania, Italy}
\affiliation{\sup{3}Department of Civil Engineering and Architecture, University of Catania, Catania, Italy}
\affiliation{\sup{4}Department of Mathematics and Computer Science, University of Catania, Catania, Italy}
\affiliation{\sup{5}School of Computing, Edinburgh Napier University, Edinburgh, UK}
\email{casella@di.unito.it, alessio.chisari@phd.unict.it, battiato@dmi.unict.it, v.giuffrida@napier.ac.uk}
}


\definecolor{Gray}{gray}{0.9}


\abstract{
It has been demonstrated that deep neural networks outperform traditional machine learning. However, deep networks lack generalisability, that is, they will not perform as good as in a new (testing) set drawn from a different distribution due to the domain shift. In order to tackle this known issue, several transfer learning approaches have been proposed, where the knowledge of a trained model is transferred into another to improve performance with different data. However, most of these approaches require additional training steps, or they suffer from catastrophic forgetting that occurs when a trained model has overwritten previously learnt knowledge. We address both problems with a novel transfer learning approach that uses network aggregation. We train dataset-specific networks together with an aggregation network in a unified framework. The loss function includes two main components: a task-specific loss (such as cross-entropy) and an aggregation loss. The proposed aggregation loss allows our model to learn how trained deep network parameters can be aggregated with an aggregation operator. We demonstrate that the proposed approach learns model aggregation at test time without any further training step, reducing the burden of transfer learning to a simple arithmetical operation. The proposed approach achieves comparable performance w.r.t. the baseline. Besides, if the aggregation operator has an inverse, we will show that our model also inherently allows for selective forgetting, i.e., the aggregated model can forget one of the datasets it was trained on, retaining information on the others.
}

\onecolumn \maketitle \normalsize \setcounter{footnote}{0} \vfill

\section{\uppercase{Introduction}}
\label{sec:introduction}

Deep Learning (DL) has demonstrated superior performance than traditional ML methods in a variety of tasks. This is due to being able to extract discriminative features from the data for the task at hand via end-to-end training. Such discriminative features are suitable for the dataset the network was trained on. However, a deep network will not perform as good as in a different dataset due to the \textit{domain shift} (or dataset bias) \cite{zhao2020}.

A way to address the domain shift is via Transfer Learning (TL), where the information learnt by a trained network is (re)used in another context. Several approaches to transfer learning have been proposed in literature, such as sample reweighting \cite{Scholkopf2007}, feature distributions minimisation \cite{Tzeng2017,Litrico2021}, distillation \cite{hinton2015distilling}, and so on (for a recent survey on TL, please read \cite{Zhuang2021}).

However, transfer learning techniques may be affected by catastrophic forgetting \cite{Goodfellow2013}, where a network forgets the information learnt from a previous task when transferred to a new one. Furthermore, generally, transfer learning requires further training steps to accommodate for new data, even though the learnt task remains unchanged.

\begin{figure*}[t]
    \centering
    \includegraphics[width=\textwidth]{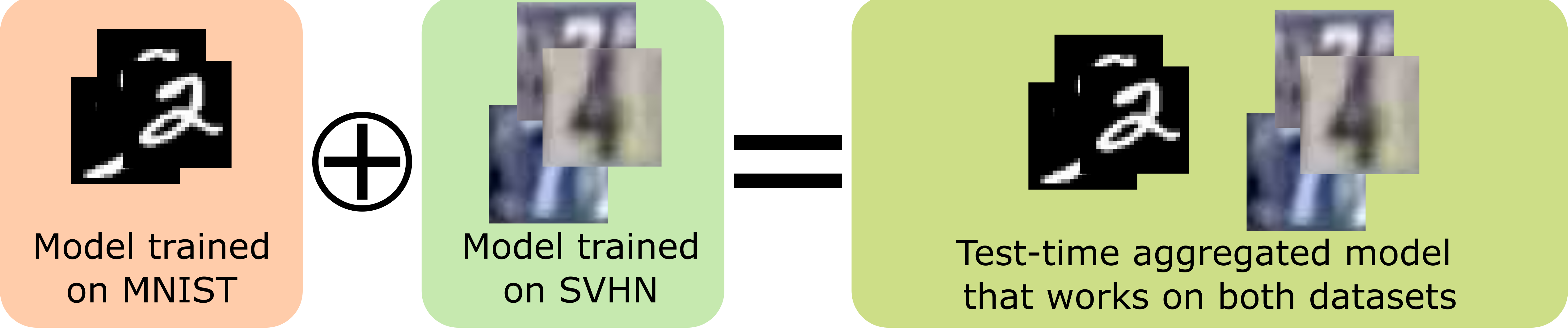}
    \caption{Pictorial representation of the proposed method that performs test-time neural network aggregation.}
    \label{fig:summary}
\end{figure*}

The benefit of transfer learning has been demonstrated extensively in the last years \cite{weiss2016}, even in distributed training scenario \cite{chen2020}. In this context, a central model is trained on several datasets that have never directly seen, as they are located in different machines (federated learning). However, this training paradigm raised another question: what if one (or more) datasets used to train the centrally trained model needs to be removed? Machine unlearning \cite{Golatkar2021} is studied for several reasons, especially when sensible data are used (e.g., medical imaging). However, it is generally hard to selectively \textit{scrub} the parameters of a model such that it cannot perform well on a portion of the dataset, whilst it retains comparable performance as before on the rest of the dataset.

In this paper, we propose a new proof-of-concept technique to TL that inherently allows for selective forgetting by aggregating the network parameters \textit{without any further training}. This can be applied to different datasets, assuming they all share the same task. Our approach is represented in \Cref{fig:summary} and works as follows: we train a VGG-like \cite{Simonyan2015} deep neural network for each dataset -- we will refer to these networks as $N_i$, for $i=1,\ldots,n$, with $n$ being the number of datasets. In addition, a VGG-like network -- named $N^*$ -- is also trained taking all the datasets as inputs. All the networks are trained end-to-end with a \textit{aggregation} regulariser, ensuring that the weights learnt by $N^*$ are obtained as an aggregation for all the other networks $N_i$. This training paradigm will ensure that the networks $N_i$ also learn how to be aggregated. Furthermore, requiring that the aggregation function is invertible, our model inherently allows for selective forgetting. In our experiments, we set $n=2$ datasets, and we used the sum of weights as network aggregation function (which can easily be inverted with subtraction), which is applied to only the parameters of the feature extractors. All the networks trained within this end-to-end framework (including $N^*$) share the same classifier. Experimental results show that test-time network aggregation is possible, outperforming the baseline.

The key contributions of our approach can be summarised as follows:

\begin{enumerate}
    \item we propose the \textit{aggregation regulariser} during training;
    \item network aggregation is achieved at test time (no further training is required);
    \item our transfer learning technique does not suffer from catastrophic forgetting;
    \item our approach can also be used for selective forgetting (assuming networks are aggregated via an invertible function).
\end{enumerate}

The rest of the paper is organised as follows. In \Cref{sec:litrev}, we discuss the recent related works. \Cref{sec:method} outlines our proposed approach. In \Cref{sec:results}, experimental results are shown and discussed. Finally, \Cref{sec:conclusion} concludes the paper.

\section{\uppercase{Related Work}}
\label{sec:litrev}

The aggregation of network parameters is a form of transfer learning. Typically, TL generally addresses a better initial and steeper growth performance \cite{Tommasi2010} by reuse of the convolutional filter parameters of CNNs. For example, fine-tuning is the simplest way to achieve transfer learning: a model, pre-trained on a dataset, e.g. ImageNet \cite{Deng2009}, is used as starting point for other datasets and tasks \cite{reyes2015fine}. Although intuitive and easy to do, fine-tuning typically underperforms wrt other transfer learning approaches \cite{shu2021,Han2021}. More sophisticated methods have been proposed \cite{Oquab2014}, but several of them suffer from \textit{negative transfer} \cite{Rosenstein05totransfer,Pan2010,Torrey2010,Wang2018,Zhuang2021}: the process of transferring knowledge is harmful because the knowledge is not transferable across all the domains (in particular when the source and target datasets are not related). 

Another issue affecting transfer learning approaches is \textit{catastrophic forgetting}, where new knowledge permanently replaces information learnt from previous tasks \cite{Goodfellow2013}. In fact, several approaches to TL, such as \textit{Batch Spectral Shrinkage} \cite{Chen2019}, attempts to solve such an issue. However, these approaches still rely on a training procedure to adapt to a new dataset (or task). However, we asked ourselves the following question: is it possible achieving TL without catastrophic forgetting at test time? We achieve that by aggregating the weights of trained networks together. 

The idea of aggregating the parameters of deep neural networks is not new in the literature. A framework that aggregates knowledge from multiple models is the \textit{Transfer-Incremental Mode Matching} (T-IMM) \cite{Geyer2019}, which enables for adaptive merging of models. It is a re-interpretation of IMM \cite{Lee2017}, a work in the context of life-long learning aiming at the sequential aggregation of models retaining good performance on all the prior tasks, rather than on transfer learning. T-IMM belongs to the field of incremental learning, a subtly different area concerning lifelong learning, in which the parameters of the $i$-th model are used as initialisation for model $i+1$. More recently, Zoo-Tuning was proposed to adaptively aggregate multiple trained models \cite{shu2021}. To achieve network aggregation, the authors proposed the \textit{AdaAgg} layer. However, this approach assumes that models are already pre-trained before being aggregated (involving a two-step learning). In our work, models are randomly initialised and then trained once end-to-end and simultaneously. 

Lifelong (or continual) learning describes the scenario in which new tasks arrive sequentially and should be incorporated into the current model, retaining previous knowledge \cite{PARISI201954}. Approaches to lifelong learning are mainly aimed to mitigate catastrophic forgetting \cite{rao2019continual,RAMAPURAM2020381,Ye2020}. According to Parisi \textit{et al.} (2019), there are three main approaches to lifelong learning: (i) retraining with regularisation; (ii) network expansion; (iii) selective network retraining and expansion. In the first case, neural networks are retrained with constraints to prevent forgetting. Network expansion approaches perform architectural changes (e.g., adding neurons) to the network to add novel information. The last approaches update only a subset of neurons and allow expansion (if necessary). Our proposed method loosely follows the paradigm of regularisation approaches with an important difference: no retraining of the architecture is performed neither transfer learning nor selective forgetting.

Our approach to network aggregation inherently allows network decomposition for selective forgetting. Recently, several related works have focused on machine unlearning \cite{Golatkar2020,Golatkar2021}. Overall, these approaches assume that the portion of the dataset that the model should unlearn is given to a \textit{scrub} function that aims to remove the information learnt from the dataset to be forgotten, impacting (although minimally) the performance of the scrubbed model on the rest of the dataset. Our approach is different: we do not require data to be provided for selective forgetting. Instead, the aggregated model trained on two (or more) datasets can be changed by applying the inverse of the aggregation function (in our case, a simple subtraction).

\section{\uppercase{Proposed Method}}
\label{sec:method}

\Cref{fig:proposed} displays the proposed approach: the general idea is to aggregate the weights of two different neural networks trained on two different datasets (sharing the same underlying task). The ultimate goal is to obtain $n$ individual networks $N_i$ such that their composition  $N_1\oplus N_2 \oplus \ldots \oplus N_i \approx N^*$ (note that the operator $\oplus$ refers to a generic network aggregation operator, which details will be provided in \Cref{sec:n_star}). Below, we will refer to an individual network $N_i$ as \textit{dataset-specific network}, whereas $N^*$ will be referred to as \textit{aggregated network}. As anticipated in \Cref{sec:introduction}, all these networks used as feature extractors share the same task network.

\subsection{Task Network}

\begin{figure}
    \centering
    \includegraphics[width=\linewidth]{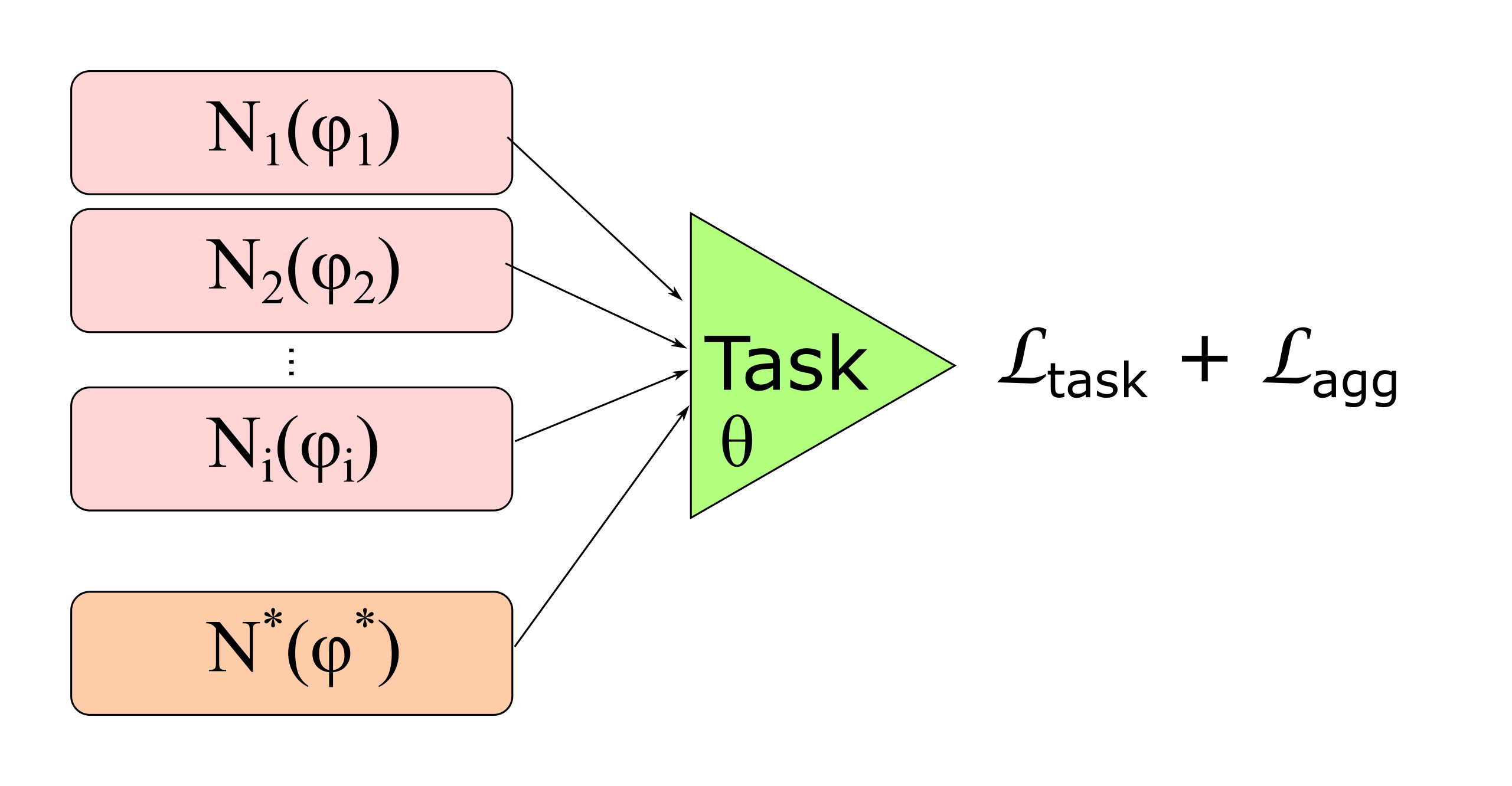}
    \caption{Graphical representation of the proposed method. Each network $N_i$ is parametrised by a set of weights $\varphi_i$. The aggregated network $N^*$ is parametrised by $\varphi^*$. There is no weight sharing between these networks. However, all the networks share the same task network (i.e., a classifier). The total objective function is given as a combination two loss functions: (i) task loss (i.e., cross-entropy); (ii) aggregation loss (see \Cref{sec:n_star}).}
    \label{fig:proposed}
\end{figure}

As shown in \Cref{fig:proposed}, the task network is shared across the aggregated and dataset-specific networks. This network is parametrised by the set of weights $\theta$: the output provided by all of the $N_i$ and $N^*$ is used as input of the task network, and its output is the prediction (i.e., softmax activation in case of classification).

The task network is trained with a task-specific loss function as $\mathcal{L}_T(z,y;w)$, that takes the training data $z$ (in form of representation) and target variables $y$ as inputs, and it is parametrised by a set of weights $w$ (that includes $\theta$ and the parameters of the feature extractors). We used cross-entropy loss for $\mathcal{L}_T$ in this paper. For other tasks (i.e., regression), a different loss function may be used (e.g., mean squared error).

\subsection{Dataset-Specific Network}
Each dataset-specific network $N_i$ acts as a feature extractor for the dataset it is trained on. We opted to use a VGG-like network \cite{Simonyan2015} in our experiments. In particular, following the same architectural tweaks as others \cite{Lho2021}, we used a VGG-16 network with Group Normalisation \cite{Wu2018}.\footnote{We also tried either Batch Normalisation \cite{ioffe2015batch} or no normalisation with no success.}

Each network $N_i$ is parametrised by the set of weights $\varphi_i$ that is trained via standard supervised learning. In fact, each $N_i$ is trained with a different label dataset; all of those datasets maintain the same underlying task. This means that each dataset $\mathcal{D}_i$ contains a set of input data $\mathcal{X}_i$, such that $x^{(i)} \in \mathcal{X}_i$, and a set of target values $\mathcal{Y}_i$, such that $y^{(i)}\in A$ (the set $A$ is a generic set defined by the task, i.e. if the task is classification, $A$ will contain all the possible classes).

For each of these networks, a specific loss function is used during training:

\begin{equation}
    \label{eq:task_i}
    \mathcal{L}_{T_i}\left(x^{(i)},y^{(i)};\phi_i \cup \theta\right) = \mathcal{L}_T \left(N_i\left(x^{(i)}\right),y^{(i)};\phi_i \cup \theta\right).
\end{equation}

\subsection{Aggregated Network}
\label{sec:n_star}

The aggregated network is similar to the dataset-specific networks: it shares the same architecture but not the weights. In fact, this network is parametrised by the set of weights $\phi^*$. 

The aggregated network is trained such that its weights can be expressed as a sum of the dataset-specific networks. To achieve this, we proposed the \textit{aggregation} regulariser. Each network $N_i$ is made of $L_i$ layers, and each layer $\ell$ is parametrised by some weights $W_i^{\ell}$$\in$$\varphi_i$.\footnote{Some types of layers, for example, convolutional layers, can be parametrised by multiple weights, such as kernels and bias. For the sake of clarity, we incorporate these weights within $W_i^{\ell}$.} Similarly, the aggregated network $N^*$ includes several layers $L$, each of those is parametrised by $W^{\ell}$. However, it is important to emphasise that all the feature extractors share the same architecture, that is, $L=L_1=L_2=\ldots=L_n$. 

During training, we want that:
\begin{equation}
\label{eq:w_sum}
W^{\ell}=W_1^{\ell}\oplus W_2^{\ell}\oplus \ldots \oplus W_n^{\ell},
\end{equation}

\noindent i.e., the weights at layer $\ell$ in the aggregated network should be equal to the aggregation of the corresponding layer weights in the dataset-specific networks. We reformulate this constrain as a regulariser during training. Assuming that $\ominus$ is the inverse operator of $\oplus$, the aggregation regulariser is then expressed as:

\begin{equation}
    \label{eq:reg}
    \mathcal{L}_{agg}(\Phi) = \sum_{\ell=1}^L W^{\ell} \ominus \left[ W_1^{\ell}\oplus W_2^{\ell}\oplus \ldots \oplus W_n^{\ell}\right],
\end{equation}

\noindent where $\Phi = \phi^* \cup \left(\bigcup_{i=1}^n\phi_i\right)$ is the set of all the weights in all the feature extractors.\footnote{Similarly as in \cite{Geyer2019}, we only aggregate weights of convolutional layers.} Although the networks learn a non-linear mapping w.r.t the task, the aggregation regulariser in \Cref{eq:reg} learns the weights $\Phi$ such that the network aggregation can be performed with a linear operation (assuming that $\oplus$ is linear).

The aggregated network takes all the input data that are used for each dataset-specific network $\mathcal{D} = \bigcup_{i=1}^n \mathcal{D}_i$ and it is trained in a supervised manner w.r.t. the task $\mathcal{L}_T$ as follows:

\begin{equation}
    \label{eq:task_star}
    \mathcal{L}_{T^*}(x,y;\phi^* \cup \theta) = \mathcal{L}_T (N^*(x),y,\phi^* \cup \theta),
\end{equation}

\noindent where $(x,y) \in \mathcal{D}$, i.e. inputs and labels are taken from all the datasets used to train the dataset-specific networks.

\subsection{Objective Function}
As shown in \Cref{fig:proposed}, the objective functions used to train our model is the following:

\begin{equation}
    \label{eq:objective}
    J(x,y;\Theta) = \mathcal{L}_{task} + \mathcal{L}_{agg},
\end{equation}

\noindent where $\Theta = \Phi \cup \theta$ is the set of all the parameters in the network. the loss function $\mathcal{L}_{task}$ is given as the sum of all the task-specific loss functions expressed in \Cref{eq:task_i} and \Cref{eq:task_star}:

\begin{equation*}
\begin{split}
    \mathcal{L}_{task}(x,y;\Theta) &= \mathcal{L}_{T^*}(x,y;\phi^* \cup \theta)\\
    &+ \sum_{i=1}^n \mathcal{L}_{T_i}\left(x^{(i)},y^{(i)};\phi_i \cup \theta\right).
\end{split}
\end{equation*}

After training, there is no guarantee that the regulariser in \Cref{eq:reg} ensures that \Cref{eq:w_sum} is satisfied. However, the optimisation of \Cref{eq:objective} will make sure that $W^{\ell}\approx W_1^{\ell}\oplus W_2^{\ell}\oplus \ldots \oplus W_n^{\ell}$. Hence, we can retrieve $N^*$ by aggregating all the weights trained for each dataset-specific network $N_i$ -- we will refer to this model as $\hat{N}^*$, such that $\hat{N}^*\approx N^*$.

We can selectively forget one of the datasets from $\hat{N}^*$ by applying a simple arithmetic operation. Assuming that we wanted to remove the $k$-th dataset, we can perform the operation $\hat{N}^* \ominus N_k$, without any further training (or adaptation) steps.

\subsection{Implementation Details}
We set the number of datasets (and thus the number of dataset-specific networks) to $n=2$. This allowed us to demonstrate whether our approach works and to set a baseline. We set the number of groups for Group Normalisation to 32. We used as aggregation operator the sum for the following reasons: (i) it has an inverse -- the subtraction; (ii) it is differentiable. We set as task-specific loss function the cross-entropy loss as we train the whole network for a classification task. Stochastic Gradient Descent (SGD) was used as optimiser for training with a learning rate $\eta=0.01$. The baseline was trained for 20 epochs, while training of our proposed method lasted 200 epochs. We implemented our approach in PyTorch \cite{Paszke2019} on Google Colaboratory.

\section{\uppercase{Experimental results}}
\label{sec:results}

\begin{figure*}[t]
    \centering
    \includegraphics[width=.9\textwidth,trim={4.5cm 0 4.5cm 0},clip]{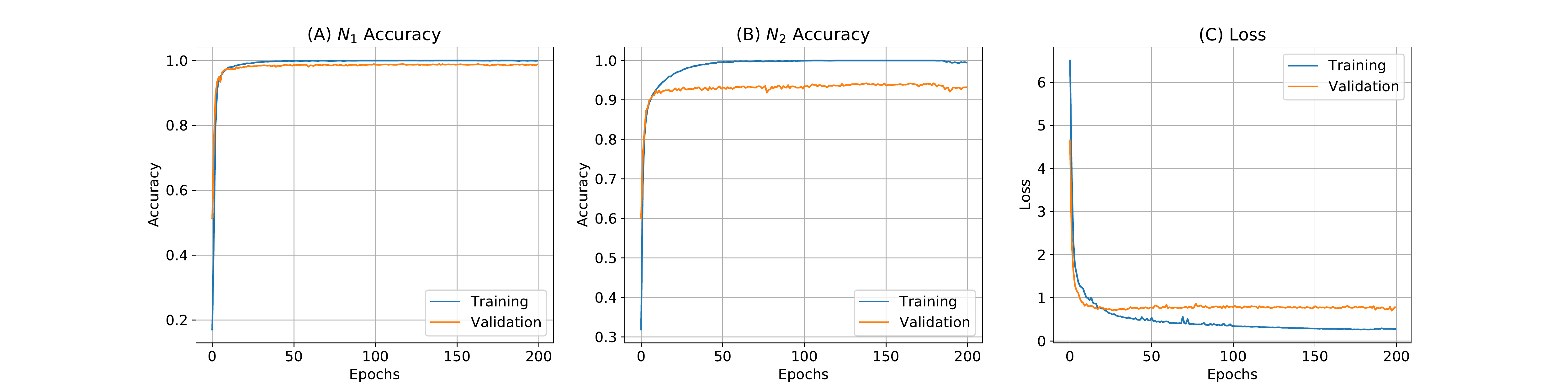}
    \caption{Training and validation accuracies and losses of our proposed method. (A) $N_1$ training and validation accuracies; (B) $N_2$ training and validation accuracies; (C) Total training and validation loss.}
    \label{fig:plots}
\end{figure*}

\mypar{Dataset} we used MNIST \cite{lecun2010} as $\mathcal{D}_1$ and SVHN format 2 \cite{Netzer2011} as $\mathcal{D}_2$. MNIST contains $60,000$ binary images of size $28\times 28$ for training and $10,000$ images for testing. SVHN contains $73,257$ colour images of size $32\times 32$ for training and $26,032$ for testing. We chose these two datasets for the following reasons: (i) they are designed for the same classification (10-class) task; (ii) data are drawn from different distributions.

\mypar{Preprocessing} 
In order to use the same architecture for both datasets, MNIST images were rescaled to $32\times 32$. We converted the SVHN images in grayscale. As for data augmentation, we performed random horizontal flips with a probability of $50\%$.

\mypar{Baseline} We compared our method with a standard VGG-16 network with Batch Normalisation \cite{ioffe2015batch}. We ran the following baseline experimentation:

\begin{enumerate}
    \item trained it only on MNIST -- following the notation adopted in this paper, we called this trained network $N_1$;
    \item trained it only on SVHN -- named $N_2$;
    \item trained it on both ($N^*$);
    \item The weights trained on $N_1$ and $N_2$ were taken to perform $N_1 \oplus N_2$ at test time.
\end{enumerate}

\noindent Experimental results are shown in \Cref{tab:results}.

\begin{table}[t]
\centering
\begin{tabular}{@{}lccccc@{}}
\toprule
                   & \multicolumn{2}{c}{Trained on} & \multicolumn{3}{c}{Tested on}                  \\

               & $\mathcal{D}_1$ & $\mathcal{D}_2$ & $\mathcal{D}_1$ & $\mathcal{D}_2$ & $\mathcal{D}_1\cup \mathcal{D}_2$ \\ \midrule
\multicolumn{6}{l}{\textit{Baseline}} \\
$N_1$      &        \checked       &      \textendash         & 99.04\%                &   8.67\%            &        33.75\%       \\
$N_2$      &     \textendash   & \checked &      57.63\%          &    90.29\%           &       80.78\%        \\
$N^*$        &   \checked     &   \checked    &      98.73\%          &      90.58\%         &     92.85\%          \\
$N_1\oplus N_2$ &     \checked   &  \checked     &   9.75\%            &     7.59\%          &      8.19\%         \\ 
\midrule
\multicolumn{6}{l}{\textit{Proposed Method}} \\

$N_1$      &        \checked        &      \textendash         &       98.90\%         &    41.45\%           &     57.35\%          \\
$N_2$      &     \textendash   & \checked  &    45.30\%            &      92.41\%         &  79.31\%             \\
$N^*$        &   \checked     &   \checked    &     
98.40\%           &     86.47\%          &   89.77\%  \\
$N_1\oplus N_2$ &     \checked   &  \checked     &    96.41\%            &     68.03\%          &     75.88\% \\
\bottomrule
\end{tabular}
\caption{Testing performance of the proposed method compared to the baseline performance. $\mathcal{D}_1$ indicates MNIST; $\mathcal{D}_2$ indicates SVHN. The models obtained via aggregation (i.e., $N_1\oplus N_2$) are obtained at test time by aggregating the weights of the networks.}
\label{tab:results}
\end{table}

\subsection{Discussion}
Our purpose is to demonstrate that the performance of our aggregated network $N_1 \oplus N_2$ is better than the baseline.  Overall, our method achieves comparable performance with the baseline for individual tasks (i.e., $N_1$ and $N_2$). However, there is a slight loss in performance in $N^*$, that is, the network trained on both MNIST and SVHN, with our method. The baseline achieves approx. $92\%$ accuracy, whereas $N^*$ trained with our method achieves approx. $89\%$.

Although this minor performance reduction, our method achieves high performance with test-time weight aggregation. After $N_1$ and $N_2$ are trained with the baseline and our method, weights are aggregated by applying the $\oplus$ operator. \Cref{tab:results} clearly shows that our training procedure outperforms the baseline (8\% vs 75\% testing accuracy). This demonstrates that a traditional training of two networks cannot be aggregated, leading to catastrophic forgetting on both datasets. Our training approach with \Cref{eq:reg} enables the networks to explicitly learn an aggregation operation that can be reproduced at test time. 

Ideally, the performance of $N_1\oplus N_2$ should be as close as possible to $N^*$. As shown in the last two lines of \Cref{tab:results}, there is an approximate loss of $14\%$ accuracy. We hypothesise several reasons for this gap in accuracy: (i) our method may require more training time; (ii) Group Normalisation may be having an impact at test time (as specified in \Cref{sec:n_star}, we only aggregate the weights of convolutional layers); (iii) use of Weight Standardisation (WS) can improve performance \cite{Qiao2019,Lho2021}. 

In relation to training time, we plot the training and validation accuracies and losses of our method in \Cref{fig:plots}. Overall, it can be noted that 50 epochs should be enough to accommodate for both datasets. However, we found experimentally that more training time results in higher performances in our aggregated model. We hypothesise that the optimisation of \Cref{eq:reg} could require more time to learn better aggregable models. In relation to Weight Standardisation, we posit that an increase in performance may be achievable if the aggregation function is changed to the mean instead of the sum. Otherwise, the aggregated weights will no longer be zero-centred.

\begin{table}[t]
\centering
\begin{tabular}{@{}lccc@{}}
\toprule

               & $\mathcal{D}_1$ &  $\mathcal{D}_2$ & $\mathcal{D}_1\cup \mathcal{D}_2$ \\ \midrule
\multicolumn{4}{l}{\textit{Commutativity}} \\

$N_1\oplus N_2$     &    96.41\%            &     68.03\%          &     75.88\%           \\ 
$N_2\oplus N_1$    &    96.41\%            &     68.03\%          &     75.88\%           \\
\midrule
\multicolumn{4}{l}{\textit{Selective forgetting}} \\ \rowcolor{Gray}
$N_1$     &   98.90\%         &    41.45\%           &     57.35\%          \\
$(N_1\oplus N_2)\ominus N_2$&    98.55\%            &     47.90\%          &     61.92\%           \\
$(N_2\oplus N_1)\ominus N_2$    &    98.55\%            &     47.90\%          &     61.92\%           \\ \rowcolor{Gray}
$N_2$      &      45.30\%            &      92.41\%         &  79.31\%      \\
$(N_1\oplus N_2)\ominus N_1$    &    34.67\%            &     90.88\%          &     75.28\%   \\
$(N_2\oplus N_1)\ominus N_1$     &    34.67\%            &     90.88\%          &     75.28\%   \\
\bottomrule
\end{tabular}
\caption{Commutativity and Selective forgetting testing results. The training is performed in both $\mathcal{D}_1$ and $\mathcal{D}_2$. The two datasets are the same as in \Cref{tab:results}. Highlighted rows are copied from \Cref{tab:results} to ease comparison.
}
\label{tab:comm_forg}
\end{table}

\subsection{Commutativity}
\label{sec:comm}
Here, we want to demonstrate whether our method is commutative: does the performance of $N_1\oplus N_2$ match the performance of $N_2\oplus N_1$? Theoretically, commutativity should be strictly related to the $\oplus$ operator. However, this is not exactly guaranteed in our framework because we only aggregate the convolution weights in the networks $N_i$ (see \Cref{sec:n_star}). Group Normalisation layers also include learnable parameters that are not included during network aggregation. To confirm whether our method is commutative, we also performed the operation $N_2\oplus N_1$ and results are reported in \Cref{tab:comm_forg}. It can be seen that the performances in both scenarios are the same. Hence, we can conclude that our approach is commutative.

\subsection{Selective Forgetting}

For the same reasons as in \Cref{sec:comm}, we also experimentally show whether selective forgetting is possible with our method. Because of Group Normalisation layers, $(N_1\oplus N_2)\ominus N_1 \approx N_2$, i.e., by removing the contribution of $N_1$, we do not exactly obtain $N_2$ (and vice versa). Therefore, we asked the following question: does the network obtained by $(N_1\oplus N_2)\ominus N_1$ perform as good as $N_2$? \Cref{tab:comm_forg} shows the experimental results of selective forgetting. 

\mypar{Forgetting SVHN} We removed the weights of $N_2$ from the aggregated networks (we considered $N_1\oplus N_2$ and $N_2\oplus N_1$ as aggregated networks), and we provided the SVHN testing set to this new network. The testing accuracy is $47.90\%$, against the $41.45\%$ of $N_1$. Therefore, the resulting network does forget about SVHN, although not completely (there is approx $+6\%$ increase of performance).

\mypar{Forgetting MNIST} A similar experiment was performed by removing the weights of $N_1$ from the aggregated networks. Differently than before, the testing accuracy of the resulting network is $34.67\%$, compared with $45.30\%$. This experimentally demonstrates that our method has completely forgotten the information learnt from the MNIST dataset.

\mypar{Retained information}  In the two previous experiments, it was shown that the resulting network had forgotten information from either of the two datasets. However, we must check whether the network can still perform well in the other dataset. Overall, the performance on MNIST dataset is very similar (from $98.90\%$ to $98.55\%$), whereas in the case of SVHN there is approx $2\%$ loss of performance (from $92.41\%$ to $90.88\%$) -- although the overall testing error is above $90\%$. This also demonstrates that the proposed method retains information from both tasks with a loss in performance up to $2\%$.

\section{\uppercase{Conclusions}}
\label{sec:conclusion}
In this paper, we proposed a novel and simple proof-of-concept transfer learning approach that inherently allows for selective forgetting. Our training method enables for network aggregation at test time, i.e. the weights of two networks (trained on two different datasets) are aggregated together, such that the resulting network can work on both datasets without any further training/adaptation step.

We achieve that by introducing an \textit{aggregation regulariser}, that enables the networks to also learn the aggregation operation in an end-to-end training framework. We used the sum as aggregation operator, as it is invertible and differentiable. VGG-like architectures were used as feature extractors, using Group Normalisation in lieu of Batch Normalisation.

Our experimental results demonstrated that the proposed approach allows for test-time transfer learning without any further training steps. Furthermore, we showed that our training procedure is commutative: the aggregated network $N_1 \oplus N_2$ obtains the same performance of $N_2 \oplus N_1$. Moreover, we demonstrated that our method allows for selective forgetting (at the cost of up to $2\%$ testing performance).

The proposed method has some limitations: (i) it requires that all the networks involved in the training share the same architecture; (ii) the selective forgetting does not allow to forget a subset of the dataset; (iii) we evaluated it on just two benchmark datasets (although the proposed framework can easily accommodate for multiple datasets). As future work, we will generalise our approach exploring the training with $N_i$ deep neural networks, for $i = 1,...,n$, with $n$ being the number of datasets, in a federated learning scenario.

\section*{\uppercase{Acknowledgements}}
This work was funded by the Edinburgh Napier University internally funded project ``Li.Ne.Co.''

\bibliographystyle{apalike}
{\small
\bibliography{example}}

\begin{thebibliography}{}

\bibitem[Chen et~al., 2019]{Chen2019}
Chen, X., Wang, S., Fu, B., Long, M., and Wang, J. (2019).
\newblock Catastrophic forgetting meets negative transfer: Batch spectral
  shrinkage for safe transfer learning.
\newblock In Wallach, H., Larochelle, H., Beygelzimer, A., d\textquotesingle
  Alch\'{e}-Buc, F., Fox, E., and Garnett, R., editors, {\em Advances in Neural
  Information Processing Systems 32}, pages 1906--1916. Curran Associates, Inc.

\bibitem[Chen et~al., 2020]{chen2020}
Chen, Y., Qin, X., Wang, J., Yu, C., and Gao, W. (2020).
\newblock Fedhealth: A federated transfer learning framework for wearable
  healthcare.
\newblock {\em IEEE Intelligent Systems}, 35(4):83--93.

\bibitem[Deng et~al., 2009]{Deng2009}
Deng, J., Dong, W., Socher, R., Li, L.-J., Li, K., and Fei-Fei, L. (2009).
\newblock Imagenet: A large-scale hierarchical image database.
\newblock In {\em 2009 IEEE Conference on Computer Vision and Pattern
  Recognition}, pages 248--255.

\bibitem[Geyer et~al., 2019]{Geyer2019}
Geyer, R., Corinzia, L., and Wegmayr, V. (2019).
\newblock Transfer learning by adaptive merging of multiple models.
\newblock In Cardoso, M.~J., Feragen, A., Glocker, B., Konukoglu, E., Oguz, I.,
  Unal, G., and Vercauteren, T., editors, {\em Proceedings of The 2nd
  International Conference on Medical Imaging with Deep Learning}, volume 102
  of {\em Proceedings of Machine Learning Research}, pages 185--196. PMLR.

\bibitem[Golatkar et~al., 2021]{Golatkar2021}
Golatkar, A., Achille, A., Ravichandran, A., Polito, M., and Soatto, S. (2021).
\newblock Mixed-privacy forgetting in deep networks.
\newblock In {\em Proceedings of the IEEE/CVF Conference on Computer Vision and
  Pattern Recognition (CVPR)}, pages 792--801.

\bibitem[Golatkar et~al., 2020]{Golatkar2020}
Golatkar, A., Achille, A., and Soatto, S. (2020).
\newblock Eternal sunshine of the spotless net: Selective forgetting in deep
  networks.
\newblock In {\em Proceedings of the IEEE/CVF Conference on Computer Vision and
  Pattern Recognition (CVPR)}.

\bibitem[Goodfellow et~al., 2013]{Goodfellow2013}
Goodfellow, I.~J., Mirza, M., Xiao, D., Courville, A., and Bengio, Y. (2013).
\newblock An empirical investigation of catastrophic forgetting in
  gradient-based neural networks.
\newblock {\em arXiv preprint arXiv:1312.6211}.

\bibitem[Han et~al., 2021]{Han2021}
Han, X., Huang, Z., An, B., and Bai, J. (2021).
\newblock Adaptive transfer learning on graph neural networks.

\bibitem[Hinton et~al., 2015]{hinton2015distilling}
Hinton, G., Vinyals, O., and Dean, J. (2015).
\newblock Distilling the knowledge in a neural network.
\newblock {\em arXiv preprint arXiv:1503.02531}.

\bibitem[Ioffe and Szegedy, 2015]{ioffe2015batch}
Ioffe, S. and Szegedy, C. (2015).
\newblock Batch normalization: Accelerating deep network training by reducing
  internal covariate shift.
\newblock In {\em International conference on machine learning}, pages
  448--456. PMLR.

\bibitem[LeCun et~al., 2010]{lecun2010}
LeCun, Y., Cortes, C., and Burges, C. (2010).
\newblock Mnist handwritten digit database.
\newblock {\em ATT Labs [Online]. Available: http://yann.lecun.com/exdb/mnist},
  2.

\bibitem[Lee et~al., 2017]{Lee2017}
Lee, S.-W., Kim, J.-H., Jun, J., Ha, J.-W., and Zhang, B.-T. (2017).
\newblock Overcoming catastrophic forgetting by incremental moment matching.
\newblock In {\em 31st Conference on Neural Information Processing Systems
  (NIPS 2017), Long Beach, CA, USA}.

\bibitem[Litrico et~al., 2021]{Litrico2021}
Litrico, M., Battiato, S., Tsaftaris, S.~A., and Giuffrida, M.~V. (2021).
\newblock Semi-supervised domain adaptation for holistic counting under label
  gap.
\newblock {\em Journal of Imaging}, 7(10).

\bibitem[Loh et~al., 2021]{Lho2021}
Loh, A., Karthikesalingam, A., Mustafa, B., Freyberg, J., Houlsby, N.,
  MacWilliams, P., Natarajan, V., Wilson, M., McKinney, S.~M., Sieniek, M.,
  Winkens, J., Liu, Y., Bui, P., Prabhakara, S., and Telang, U. (2021).
\newblock Supervised transfer learning at scale for medical imaging.

\bibitem[Netzer et~al., 2011]{Netzer2011}
Netzer, Y., Wang, T., Coates, A., Bissacco, A., Wu, B., and Ng, A.~Y. (2011).
\newblock Reading digits in natural images with unsupervised feature learning.
\newblock In {\em NIPS Workshop on Deep Learning and Unsupervised Feature
  Learning 2011}.

\bibitem[Oquab et~al., 2014]{Oquab2014}
Oquab, M., Bottou, L., Laptev, I., and Sivic, J. (2014).
\newblock Learning and transferring mid-level image representations using
  convolutional neural networks.
\newblock In {\em Proceedings of the IEEE conference on computer vision and
  pattern recognition}, page pages 1717–1724.

\bibitem[Pan and Yang, 2010]{Pan2010}
Pan, S.~J. and Yang, Q. (2010).
\newblock A survey on transfer learning.
\newblock {\em IEEE Transactions on Knowledge and Data Engineering}, 22.

\bibitem[Parisi et~al., 2019]{PARISI201954}
Parisi, G.~I., Kemker, R., Part, J.~L., Kanan, C., and Wermter, S. (2019).
\newblock Continual lifelong learning with neural networks: A review.
\newblock {\em Neural Networks}, 113:54--71.

\bibitem[Paszke et~al., 2019]{Paszke2019}
Paszke, A., Gross, S., Massa, F., Lerer, A., Bradbury, J., Chanan, G., Killeen,
  T., Lin, Z., Gimelshein, N., Antiga, L., Desmaison, A., Kopf, A., Yang, E.,
  DeVito, Z., Raison, M., Tejani, A., Chilamkurthy, S., Steiner, B., Fang, L.,
  Bai, J., and Chintala, S. (2019).
\newblock Pytorch: An imperative style, high-performance deep learning library.
\newblock In {\em Advances in Neural Information Processing Systems 32}, pages
  8024--8035. Curran Associates, Inc.

\bibitem[Qiao et~al., 2019]{Qiao2019}
Qiao, S., Wang, H., Liu, C., Shen, W., and Yuille, A. (2019).
\newblock Micro-batch training with batch-channel normalization and weight
  standardization.

\bibitem[Ramapuram et~al., 2020]{RAMAPURAM2020381}
Ramapuram, J., Gregorova, M., and Kalousis, A. (2020).
\newblock Lifelong generative modeling.
\newblock {\em Neurocomputing}, 404:381--400.

\bibitem[Rao et~al., 2019]{rao2019continual}
Rao, D., Visin, F., Rusu, A.~A., Teh, Y.~W., Pascanu, R., and Hadsell, R.
  (2019).
\newblock Continual unsupervised representation learning.
\newblock {\em arXiv preprint arXiv:1910.14481}.

\bibitem[Reyes et~al., 2015]{reyes2015fine}
Reyes, A.~K., Caicedo, J.~C., and Camargo, J.~E. (2015).
\newblock Fine-tuning deep convolutional networks for plant recognition.
\newblock {\em CLEF (Working Notes)}, 1391:467--475.

\bibitem[Rosenstein et~al., 2005]{Rosenstein05totransfer}
Rosenstein, M.~T., Marx, Z., Kaelbling, L.~P., and Dietterich, T.~G. (2005).
\newblock To transfer or not to transfer.
\newblock In {\em In NIPS’05 Workshop, Inductive Transfer: 10 Years Later}.

\bibitem[Schölkopf et~al., 2007]{Scholkopf2007}
Schölkopf, B., Platt, J., and Hofmann, T. (2007).
\newblock {\em Correcting Sample Selection Bias by Unlabeled Data}, pages
  601--608.

\bibitem[Shu et~al., 2021]{shu2021}
Shu, Y., Kou, Z., Cao, Z., Wang, J., and Long, M. (2021).
\newblock Zoo-tuning: Adaptive transfer from a zoo of models.
\newblock In Meila, M. and Zhang, T., editors, {\em Proceedings of the 38th
  International Conference on Machine Learning}, volume 139 of {\em Proceedings
  of Machine Learning Research}, pages 9626--9637. PMLR.

\bibitem[Simonyan and Zisserman, 2015]{Simonyan2015}
Simonyan, K. and Zisserman, A. (2015).
\newblock Very deep convolutional networks for large-scale image recognition.
\newblock {\em ICLR 2015}.

\bibitem[Tommasi et~al., 2010]{Tommasi2010}
Tommasi, T., Orabona, F., and Caputo, B. (2010).
\newblock Safety in numbers: Learning categories from few examples with multi
  model knowledge transfer.
\newblock In {\em roceedings of IEEE Computer Vision and Pattern Recognition
  Conference}.

\bibitem[Torrey and Shavlik, 2010]{Torrey2010}
Torrey, L. and Shavlik, J. (2010).
\newblock Transfer learning.
\newblock {\em Handbook of Research on Machine Learning Applications and
  Trends}.

\bibitem[Tzeng et~al., 2017]{Tzeng2017}
Tzeng, E., Hoffman, J., Saenko, K., and Darrell, T. (2017).
\newblock Adversarial discriminative domain adaptation.
\newblock In {\em Proceedings of the IEEE Conference on Computer Vision and
  Pattern Recognition (CVPR)}.

\bibitem[Wang et~al., 2018]{Wang2018}
Wang, Z., Dai, Z., Póczos, B., and Carbonell, J. (2018).
\newblock Characterizing and avoiding negative transfer.

\bibitem[Weiss et~al., 2016]{weiss2016}
Weiss, K., Khoshgoftaar, T.~M., and Wang, D. (2016).
\newblock A survey of transfer learning.
\newblock {\em Journal of Big data}, 3(1):1--40.

\bibitem[Wu and He, 2018]{Wu2018}
Wu, Y. and He, K. (2018).
\newblock Group normalization.
\newblock In {\em Proceedings of the European conference on computer vision
  (ECCV)}, pages 3--19.

\bibitem[Ye and Bors, 2020]{Ye2020}
Ye, F. and Bors, A.~G. (2020).
\newblock Learning latent representations across multiple data domains using
  lifelong vaegan.
\newblock In Vedaldi, A., Bischof, H., Brox, T., and Frahm, J.-M., editors,
  {\em Computer Vision -- ECCV 2020}, pages 777--795, Cham. Springer
  International Publishing.

\bibitem[Zhao et~al., 2020]{zhao2020}
Zhao, S., Yue, X., Zhang, S., Li, B., Zhao, H., Wu, B., Krishna, R., Gonzalez,
  J.~E., Sangiovanni-Vincentelli, A.~L., Seshia, S.~A., et~al. (2020).
\newblock A review of single-source deep unsupervised visual domain adaptation.
\newblock {\em IEEE Transactions on Neural Networks and Learning Systems}.

\bibitem[Zhuang et~al., 2021]{Zhuang2021}
Zhuang, F., Qi, Z., Duan, K., Xi, D., Zhu, Y., Zhu, H., Xiong, H., and He, Q.
  (2021).
\newblock A comprehensive survey on transfer learning.
\newblock {\em Proceedings of the IEEE}, 109(1):43--76.

\end{thebibliography}

\end{document}